\documentstyle[icassp91]{article}

\title{Language Modelling For Task-Oriented Domains}

\name{\parbox{2.7in}{\centering Cosmin Popovici}
\ ~~~~~~~~ \
\parbox{3in}{\centering Paolo Baggia}}

\address{
\parbox[t]{2.7in}{\centering {\small ICI -
Institutul de Cercetari in Informatica\\
Bd. M. Averescu, 8-10\\
Bucuresti (Romania)}}
\ ~~~~~~~~ \
\parbox[t]{3in}{\centering {\small CSELT -
Centro Studi e Laboratori Telecomunicazioni\\
Via G. Reiss Romoli, 274\\
I-10148 Torino (Italy)\\
{\tt baggia@cselt.stet.it}}}}

\begin{document}
\input{epsf}
\maketitle
\bibliographystyle{fullname}

\begin{abstract}
This paper is focused on the language modelling for 
task-oriented domains and presents an accurate analysis 
of the utterances acquired by the Dialogos spoken 
dialogue system. Dialogos allows access to the Italian 
Railways time\-table by using the telephone over the 
public network.

The language modelling aspects of specificity and 
behaviour to rare events are studied. A technique for 
getting a language model more robust, based on 
sentences generated by grammars, is presented. 
Experimental results show the benefit of the proposed 
technique. The increment of performance between 
language models created using grammars and usual ones, 
is higher when the amount of training material is limited. 
Therefore this technique can give an advantage 
especially for the development of language models in a 
new domain.
\end{abstract}

\section{INTRODUCTION}
Statistical language modelling (LM) is currently used 
for two different classes of applications: dictation 
systems and task-oriented spoken dialogue systems 
(SDS). 

The first kind of systems are tested with a very large 
vocabulary (60-20,000 words) and they need the 
availability of a huge amount of training data, for 
instance WSJ-NAB has a 45 million word text
corpora~\cite{cit8}.

SDSs are used in specific task-oriented domains, and 
they need special training material, which can be 
obtained either by expensive simulations~\cite{cit6} or
by using 
the SDS itself. The use of a general task-independent 
corpus for LM of a SDS could increase, in comparison to 
LM that use a task-dependent one, the perplexity by an 
order of magnitude~\cite{cit9}.
This is due to the mismatch 
between the general corpus and the specific application 
domain. In any case the acquired material is very limited, 
for instance the LM in the Air Travel Information System 
(ATIS) is based on a training-set of only 250,000
words~\cite{cit10}.

This paper is focused on the language modelling for 
task-oriented domains. The tests made uses the utterances 
acquired by the Dialogos, the SDS which allows access 
to the Italian Railways timetable by using the telephone 
over the public network~\cite{cit1}. Other similar systems are 
described in~\cite{cit2,cit5,cit7}.
The vocabulary of Dialogos contains 3,471 words, 
clustered in 358 classes. The semantically important 
words are grouped into classes, such as city names (2,983 
words), numbers (76 words), and so on. During the 
recognition, a class-based bigram LM is used, and the 
25-best sequences are rescored using a trigram LM.

Section~\ref{sec2} shows how well a LM captures the 
specificity of the domain, while Section~\ref{sec3} studies
the behaviour of the LM to rare events. Finally Section~\ref{sec4} 
illustrates a technique for generalising a LM by adding
n-grams generated by a grammar.

\section{SPECIFICITY OF A LANGUAGE MODEL}
\label{sec2}

A relevant characteristic of a task-oriented domain is 
the distribution of the user utterances in a corpus. Using 
the Dialogos SDS, a corpus of 1,363 spoken dialogues 
has been acquired, from 493 unexperienced subjects, that 
called the system from all over Italy~\cite{cit1,cit3}.

For the present study, the collected material was 
divided into two parts: a training-set of 20,511 utterances 
and a test-set of 2,040 utterances. Each utterance was 
transformed in a normalised form (NU), by changing 
each city name, month name and number into a class tag. 
For instance the user utterance:

{\em ``I want to leave from Naples to Rome Monday at five 
(o'clock)''}

becomes the following NU:

{\em ``I want to leave from CITY-NAME to CITY-NAME 
WEEK-DAY at HOUR-NUMBER''.}

For the sake of the language modelling, the NU is 
equivalent to the original utterance~\footnote{This is
because these classes are being used by the 
class-based LM and each word in a class has been 
considered with equal probability.}.

\begin{figure}[htb]
\epsfbox[100 598 338 720]{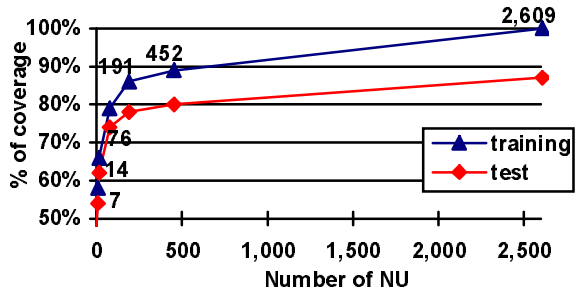}
\vspace{-5mm} 
\caption{\label{fig1} Coverage of training and test sets by the NUs.}
\end{figure}

It is worth noticing that even a small number of very 
frequent NUs cover a great part of the acquired data (see 
Figure~\ref{fig1}). The 7-th most frequent NUs cover 58\% of the 
training-set, and 54\% of test-set, and the first 191-st 
cover nearly 80\% of test-set and over 85\% of training-
set. On the other hand the NUs with just one occurrence 
are 2,060, and more then 56\% of them contain some 
spontaneous speech phenomena. This result shows that a 
few frequent NUs can already give a quite sensible 
picture of the user utterance distribution. 

\begin{figure}[htb]
\epsfbox[110 600 336 712]{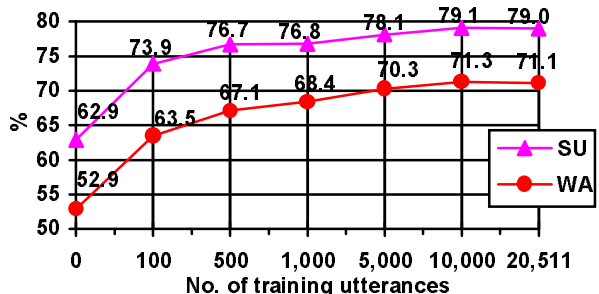}
\vspace{-3mm} 
\caption{\label{fig2} Variation of performance with size of training data.}
\end{figure}

Moreover some partial training-sets were selected, which 
include the first n utterances in the whole training set, for 
n ranging from 100 to 20,511 utterances. For each partial 
training-set a LM was created and the recognition (WA) 
and understanding (SU) rates are given in Figure~\ref{fig2}. The 
performances of the LMs created on a partial training-set 
were compared with an experiment without any LM, 
which is even reported in Figure~\ref{fig2} as 0-utterance 
training-set. A LM trained on only 100 utterances 
achieves a remarkable error rate reduction of 30\% of SU 
and 23\% of WA, especially if it is compared with the 
error reduction when the whole 20,511 training-set is 
used, that is of 43\% of SU and 39\% of WA.
 
\begin{figure}
\epsfbox[98 587 325 710]{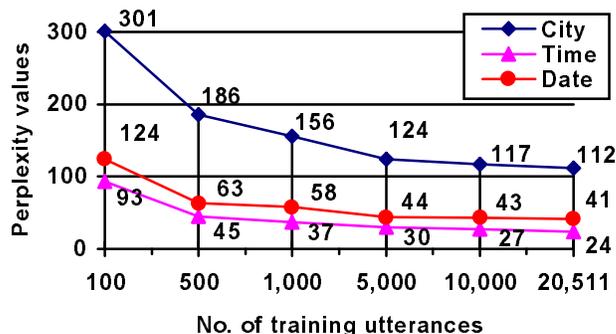}
\vspace{-4mm} 
\caption{\label{fig3} Variation of perplexity with size of training data.}
\end{figure}

A coherent behaviour is also confirmed by perplexity 
values (PP) depicted in Figure~\ref{fig3}, where the utterances 
were classified according to the kind of prompt generated 
by the system. Three representative points have been 
selected, which are the request of: departure and arrival 
city (City), time of departure (Time), and date of 
departure (Date). For these categories the PP of a 
100-utterance LM is two times higher than a 1,000-utterance 
one and three times the LM trained on the whole 
training-set. The fact that, the PP values for the City 
requests are the highest, can be explained by the large 
number of city-names in the vocabulary (2,983, near 85\% 
of the whole vocabulary).
 
\section{ROBUSTNESS TO RARE EVENTS}
\label{sec3}

In this Section the behaviour of the LM with respect 
to rare events is studied. The test-set of 2,040 utterances 
was split into two parts: The first part contains 362 
utterances, whose 351 NUs do not appear in any of the 
partial training-sets. This is referred below as the unseen 
part of the test-set. The second part includes the rest of 
the test-set (1,678 utterances, but only 257 NUs). The 
NUs in the partial training-sets cover progressively the 
utterances of the second part. For instance, the 
100-utterance training-set contains only 29 NUs, which cover 
1,317 of these 1,678 utterances. 

\begin{figure}[b]
\epsfbox[110 627 336 715]{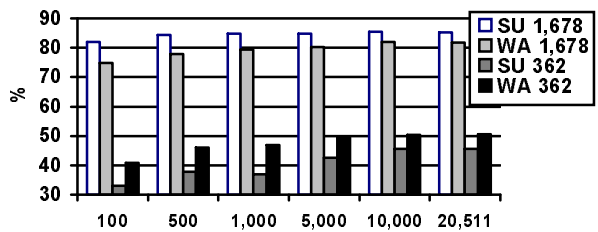}
\vspace{-4mm} 
\caption{\label{fig4} Evaluation of trained and untrained part
of the test-DB.}
\end{figure}

Both recognition, and overall understanding results 
show quite similar values for the 1,678 utterances 
(82-85\% of SU), but they are very different for the unseen 
part (33-46\% of SU), see Figure~\ref{fig4}. The performance on 
the unseen part is an indicator of the robustness of the 
model. In the following the reason for the low 
performance on the unseen part is further analysed.
 
The NUs with more then three occurrences in the 
global training-set, and different one to each other, were 
selected. Table~\ref{tab1} shows the number of this NUs, that 
exists in each one of the partial training-sets. They were 
divided into groups according to the different kind 
system request. The growth of NUs for City and Date is 
fast until 5,000 utterances are reached, then it becomes 
very slow. This indicates that there is a kind of 
saturation. While Time NUs increase nearly 
proportionally. 
 
\begin{table}[htb]
\begin{center}
\begin{small}
\begin{tabular}{|l|c|c|c|c|c|c|} \hline
            & \multicolumn{6}{c|}{training utterances}    \\ \hline
            & 100 & 500 & 1,000 & 2,000 & 10,000 & 20,511 \\ \hline
{\bf City}  &  11 &  17 &  26   &  43   &  46    &  47    \\ \hline
{\bf Date}  &   8 &  17 &  25   &  43   &  48    &  51    \\ \hline
{\bf Time}  &   6 &  12 &  15   &  36   &  42    &  49    \\ \hline
\end{tabular}
\end{small}
\end{center}
\vspace{-3mm} 
\caption{\label{tab1} Number of frequent NUs in partial training-sets.}
\end{table}

Moreover, the NUs, whose frequency in the
training-set is greater than 0.1\%, were compared with the ones in 
the test-set. We observed that the selected NUs of the 
training-set covers more then 90\% of the test-set NUs, in 
case of City and Date, but only 55\% in case of Time. 
Therefore, the City and the Date groups are considered 
much more robust than the Time group, because the 
frequent NUs do not indicate a saturation, and because 
there is a lack of the training-set NUs in the test-set. This 
is due to the high variability of the time expressions.

\section{INCREASING ROBUSTNESS BY ADDING
N-GRAMS GENERATED BY GRAMMARS}
\label{sec4}

Another coverage test was made using grammars. A 
grammar was created (explained in Section~\ref{sec4-1}) on the 
basis of the NUs in the 500-utterance training-set. The 
sentences generated by the grammar showed a coverage 
of 85\% of the NUs in the 20,511 training-set. This 
suggests that, the robustness of a LM may be increased 
by the use of a simple grammar derived from the 
common NUs in the training material.

At first the sentences generated by grammar were 
added to the training material. The obtained LMs, did not 
improve results, because the addition of the grammar 
generated sentences, greatly changes the frequency 
distribution of the n-grams, and reduces the specificity of 
the training-set.

The adopted solution was to create the LM starting 
from a data-base that contains n-grams, and not from a 
data-base of generated sentences. This made possible to 
add only the not-existing n-grams which do not highly 
affect the specificity. Therefore the tool used for training 
the LMs was changed, in order to be able to process both 
sentences and n-grams. Commonly when the n-grams are 
extracted from a sentence, they get automatically all their 
contexts (the (n-1)-gram that precedes the n-th word of 
the n-gram). On the other hand, if an n-gram is 
artificially added, it is necessary to incorporate even the 
missing contexts for this n-gram.

\subsection{Grammar creation}
\label{sec4-1}

The grammars used in the following tests were 
manually created, and they started from a set of correct 
NUs selected from a training-set. For each NU, semantic 
concepts were identified, then for each of these concepts 
a non-terminal was introduced, and, finally, each
non-terminal was generalised. For instance, in the case of a 
{\bf Time} NU:

{\em ``in the morning after seven o'clock''}, 

the following non-terminal sequence could be identified:

 {\em Part\_of\_Day Time\_Specifier Time\_Identifier}. 

{\em Part\_of\_Day} can become also {\em ``in the afternoon''},
{\em ``in the evening''} or {\em ``at lunch time''}, \newline
{\em Time\_Specifier} can be 
expressed as: {\em ``before''}, {\em ``not earlier than''},
while for {\em Time\_Identifier} other forms are: 
{\em ``a quarter to seven''}, {\em ``twenty minutes past seven''}.

At this point both the 1,000-utterance training-set (SPTS-1,000)
 and the global one (STS) were split 
according to the system request. Concentrating the 
analysis on the {\bf City}, {\bf Date}, and {\bf Time}
 requests, for the 
syntactically and semantically correct NUs in
SPTS-1,000 a grammar was created. For instance, there are 107 
NUs in the SPTS-1,000 {\bf Date} requests, and 2,483 NUs in 
STS.

For {\bf Date} and {\bf Time} requests group one grammar was 
created (Gr\_D, and Gr\_T respectively), whereas two for 
the {\bf City} requests: Gr\_C which generalises only NUs 
about departure and arrival location, and Gr\_Cdt which 
also generalises data and time, because the answers to the 
{\bf City} requests could also contain that information. 

\begin{table}[htb]
\begin{center}
\begin{footnotesize}
\begin{tabular}{|l|l|c|c|c|c|c|c|c|c|} \hline
\multicolumn{1}{|c|}{request} & \multicolumn{1}{|c|}{gram.} 
            & \multicolumn{2}{c|}{used}    & \multicolumn{2}{c|}{rare}
            & \multicolumn{2}{c|}{unknown} & \multicolumn{2}{c|}{BaFa}      
\\\hline
\multicolumn{1}{|c|}{group} & \multicolumn{1}{|c|}{used} 
                      & part & all   & part & all & part  & all & part & all
\\ \hline
{\bf City}  & Gr\_C   & 534 &  9,861 & 159  & 859 &   166 &    43 &  1 &  1 
\\ \hline
{\bf City}  & Gr\_Cdt & 568 & 10,088 & 125  & 632 & 1,326 & 1,156 &  4 &  2 
\\ \hline
{\bf Date}  & Gr\_D   & 316 &  6,921 &  96  & 849 &   150 &    82 &  1 &  1 
\\ \hline
{\bf Time}  & Gr\_T   & 276 &  6,431 &  36  & 616 & 1,748 & 1,488 & 10 &  1 
\\ \hline
\end{tabular}
\end{footnotesize}
\end{center}
\vspace{-3mm} 
\caption{\label{tab2} Event composition of the training-sets.}
\end{table}

\subsection{Creation of generalised LMs}
\label{sec4-2}

The merge between the n-grams extracted from a 
training set and from sentences generated by the 
grammar was done using the following technique. At 
first, both the training-set and the sentences generated by 
a grammar were transformed in n-grams (n=3), then three 
type of events were considered: n-grams which are 
present both in the training-set and in the generated 
sentences (called {\em usual events}), n-grams which exist only 
in the training-set (called {\em rare events}), and n-grams 
which exist only in the generated sentences (called 
{\em unknown events}). 

Into the new LM, the unknown events were added 
only once, while the rare events maintained their 
frequencies (which is quite low). In many cases the 
number of unknown events is much more higher than the 
number of usual events. For instance in the case of time 
there are 276 usual events obtained from SPTS-1,000, 36 
rare events and 1,748 unknown events. Therefore the 
quantities of usual and unknown events are weighted, by 
multiplying them with a balance-factor. At this point, a 
language model is created, then the best value for the 
balance-factor (BaFa) is empirically determined by the 
minimisation of the PP on the test-set.
 
Using Table~\ref{tab2}, the event composition of each one of 
the studied LMs can be computed. For each request 
group many LMs were created by the generalisation of 
SPTS-1,000 and STS, respectively {\em part} and {\em all} in the 
Table. It is worth noticing that in a baseline LM only the 
usual and rare events are considered.

\subsection{Experimental Results}
\label{sec40}

In this Section, the performances of the LMs that 
include n-grams generated by a grammar were compared 
with baseline LMs which does not make use of grammar 
n-grams. These baseline LMs are reported in the
Tables~\ref{tab3}-\ref{tab6}, with the tag {\em unused} in the
grammar column.
 
\begin{table}[htb]
\begin{center}
\begin{small}
\begin{tabular}{|l|l|c|c|c|c|} \hline
\multicolumn{1}{|c|}{request} & \multicolumn{1}{c|}{grammar} 
  & \multicolumn{2}{c|}{SPTS-1000} & \multicolumn{2}{c|}{STS}  \\ \hline
\multicolumn{1}{|c|}{groups}  & \multicolumn{1}{c|}{used} 
                           & WA   & SU   & WA   & SU    \\ \hline
{\bf City}  & {\em unused} & 77.5 & 68.5 & 82.3 & 71.4  \\ \hline
{\bf City}  & Gr\_C        & 78.8 & 69.3 & 82.5 & 71.4  \\ \hline
{\bf City}  & Gr\_Cdt      & 80.0 & 70.1 & 82.3 & 72.6  \\ \hline
{\bf Date}  & {\em unused} & 82.0 & 80.9 & 82.8 & 80.9  \\ \hline
{\bf Date}  & Gr\_D        & 82.7 & 81.3 & 82.9 & 80.9  \\ \hline
{\bf Time}  & {\em unused} & 79.7 & 85.5 & 83.6 & 86.7  \\ \hline
{\bf Time}  & Gr\_T        & 82.5 & 86.1 & 83.6 & 86.7  \\ \hline
\end{tabular}
\end{small}
\end{center}
\vspace{-3mm} 
\caption{\label{tab3}Recognition and understanding results.}
\end{table}

Table~\ref{tab3} shows that the LMs created using the 
grammars, obtain better results for the SPTS-1,000 LMs, 
while for the STS LMs the increment is rather limited. In 
particular, for {\bf Time} and {\bf City} the improvement of WA is 
significant. The reasons are: the high variability of time 
expressions and the fact that sometimes the {\bf City} requests 
even include information about {\bf Date} and {\bf Time}, especially 
in the first utterance to the system. This fact is evident 
from the improvement obtained by the use of the Gr\_Cdt 
grammar, which even increases the performance of the 
STS LM. 

Moreover the merge of with SPTS-1,000 with 
grammars improve the results, but they could not reach 
the performances of the baseline STS LMs. An 
explanation is that the used grammars do not model the 
highly frequent extra-linguistic phenomena.

In addition the perplexity of these LMs has been 
studied. For each group the analyses of the PP has been 
performed on the test-set and even on the sentences 
generated by the grammar. Table~\ref{tab4} shows PP results for 
all the LMs tested on the specific part of the test-set. The 
generalisation of the LMs by using grammar n-grams 
does not significantly affect the PP.
 
\begin{table}[htb]
\begin{center}
\begin{small}
\begin{tabular}{|l|l|c|c|} \hline
\multicolumn{1}{|c|}{request} & \multicolumn{1}{c|}{grammar} 
                      & SPTS-1000 & STS  \\ \hline
\multicolumn{1}{|c|}{groups}  & \multicolumn{1}{c|}{used} 
                           & PP   & PP   \\ \hline
{\bf City}  & {\em unused} &  117 & 79   \\ \hline
{\bf City}  & Gr\_C        &  118 & 78   \\ \hline
{\bf City}  & Gr\_Cdt      &  122 & 96   \\ \hline
{\bf Date}  & {\em unused} &   33 & 24   \\ \hline
{\bf Date}  & Gr\_D        &   32 & 24   \\ \hline
{\bf Time}  & {\em unused} &   20 & 14   \\ \hline
{\bf Time}  & Gr\_T        &   19 & 15   \\ \hline
\end{tabular}
\end{small}
\end{center}
\vspace{-3mm} 
\caption{\label{tab4}Perplexity results on the test-set.}
\end{table}

The use of a test-set of sentences generated by the 
grammars, even if it does not give a correct insight of the 
behaviour of the system on a test-set acquired from real 
users, because the sentence distribution is artificial, it can 
show the degree of generalisation. These PP results have 
been reported in Table~\ref{tab5} and Table~\ref{tab6} 
according to the 
number of unknown events reported in Table~\ref{tab2}. In the 
former are shown the results for small grammars (G\_C, 
and G\_D), while in the latter the results for large ones 
(G\_Cdt, and G\_T).
 
\begin{table}[htb]
\begin{center}
\begin{small}
\begin{tabular}{|l|l|c|c|} \hline
\multicolumn{1}{|c|}{request} & \multicolumn{1}{c|}{grammar} 
                     & SPTS-1000 & STS  \\ \hline
\multicolumn{1}{|c|}{groups}  & \multicolumn{1}{c|}{used} 
                           & PP  & PP   \\ \hline
{\bf City}  & {\em unused} &  72 & 36   \\ \hline
{\bf City}  & Gr\_C        &  24 & 24   \\ \hline
{\bf Date}  & {\em unused} &  52 & 25   \\ \hline
{\bf Date}  & Gr\_D        &  17 & 16   \\ \hline
\end{tabular}
\end{small}
\end{center}
\vspace{-3mm} 
\caption{\label{tab5}Perplexity results on the grammar sentences.}
\end{table}
\begin{table}[htb]
\begin{center}
\begin{small}
\begin{tabular}{|l|l|c|c|} \hline
\multicolumn{1}{|c|}{request} & \multicolumn{1}{c|}{grammar} 
                      & SPTS-1000 & STS  \\ \hline
\multicolumn{1}{|c|}{groups}  & \multicolumn{1}{c|}{used} 
                           & PP   & PP   \\ \hline
{\bf City}  & {\em unused} &  117 & 207  \\ \hline
{\bf City}  & Gr\_Cdt      &   36 &  42  \\ \hline
{\bf Time}  & {\em unused} &  231 &  53  \\ \hline
{\bf Time}  & Gr\_T        &   12 &  11  \\ \hline
\end{tabular}
\end{small}
\end{center}
\vspace{-3mm} 
\caption{\label{tab6}Perplexity values for Gr\_C and Gr\_T.}
\end{table}

In Table~\ref{tab5}, a clear reduction of the PP could be 
observed for the LMs which includes grammar n-grams. 
This reduction is higher for the LMs trained over SPTS-
1,000 (66\%), but it is relevant even for the LMs trained 
on STS (33\%).

Making a similar comparison of the PP results, 
presented in Table~\ref{tab6}, for the large sets of unknown evens, 
as expected, a more significant reduction was obtained, 
that goes from a minimum of 77\% to a maximum of 
94\%.

\section{\label{sec5}CONCLUSIONS}
This papers shows that, in a task-oriented domain, a 
LM trained out with a small amount of training material 
(1,000 utterances) acquired form naive users, allows to 
obtain rather good results, especially in the case of the 
more common NUs. This is because common NUs are a 
few, but very frequent.

Secondly, in a task-oriented domain with a very 
limited training-set, the robustness of a language 
modelling can be increased by the use of a simple 
grammar derived from the common NUs in the training 
material.

A technique for the generalisation of a language 
model adding n-grams generated by a grammar is 
described. The advantage of this technique is shown by 
experimental results. The improvements obtained by 
using this technique, are especially good for language 
models trained on a small amount of training material, 
and therefore the technique can be used in the first 
phases of the development of a LM for a new domain. 
Even if the generalised LMs do not increase the 
performance of a model trained on a large training-set, 
the perplexity indicates a better behaviour of the models 
in the case of rare events.


\begin{thebibliography}{}

\vspace{-1mm}
\bibitem{cit1}
Albesano, D., P.~Baggia, M.~Danieli, R.~Gemello,
E.~Gerbino, C.~Rullent, ``Dialogos: A Robust System for 
Human-Machine Spoken Dialogue on the Telephone'', in 
{\em Proc. of ICASSP'97}, München, 1997, vol.~2, pp.~1147--1150.

\vspace{-1mm}
\bibitem{cit2}
Aust, H., M.~Oerder, F.~Seide, V.~Steinbiss, ``The Philips 
Automatic Train Timetable Information System'', in 
{\em Speech Communications}, 1995, vol.~17, pp.~249--262.

\vspace{-1mm}
\bibitem{cit3}
Baggia, P., E.~Gerbino, E.~Giachin, C.~Rullent, 
``Experiences of Spontaneous Speech Interaction with a 
Dialogue System'', in {\em Proc. of CRIM/FORWISS 
Workshop}, München, 1994, pp.~241--248.

\vspace{-1mm}
\bibitem{cit4}
Besling,~S., H.-G.~Meier, ``Language Model Speaker 
Adaptation'', in {\em Proc. of EUROSPEECH'95}, Madrid, 
1995, pp.~1755--1758.

\vspace{-1mm}
\bibitem{cit5}
Eckert, W., T.~Kuhn, H.~Niemann, S.~Rieck, A.~Scheuer, 
E.~G.~Schukat-Talamazzini, ``A Spoken Dialogue System 
for German Intercity Train Timetable Inquiries'', in {\em Proc. 
of EUROSPEECH'93}, Berlin, 1993, vol.~3, pp.~1871--1874.

\vspace{-1mm}
\bibitem{cit6}
Fraser, N., G.~N.~Gilbert, ``Simulating Speech Systems'', in 
{\em Computer Speech and Language}, 1991, vol.~5, pp.~81--99.

\vspace{-1mm}
\bibitem{cit7}
Goddeau, D., E.~Brill, J.~Glass, C.~Pao, M.~Phillips,
J.~Polifroni, S.~Seneff, V.~Zue, ``GALAXY: A Human 
Language Interface to On-line Travel Information'', in 
{\em Proc. of ICSLP'94}, Yokoama, 1994, pp.~707--710.

\vspace{-1mm}
\bibitem{cit8}
Gauvain, J.~L., L.~Lamel, G.~Adda, D.~Matrouf, 
``Developments in Continuous Speech Dictation using the 
1995 ARPA NAB New Task'', in {\em Proc. of ICASSP'96}, 
Atlanta, 1996, vol.~1, pp.~73--76.

\vspace{-1mm}
\bibitem{cit9}
Placeway, P., R.~Schwartz, P.~Fung, L.~Nguyen, ``The 
Estimation of Powerful Language Models from Small 
and Large Corpora'', in {\em Proc. of ICASSP'93}, 
Minneapolis, 1993, vol.~2, pp.~33--36.

\vspace{-1mm}
\bibitem{cit10}
Ward, W., S.~Issar, ``Recent Improvements in the CMU 
Spoken Language Understanding System'', in {\em Proc. of 
ARPA HLT Workshop}, March 1994, pp.~213--216.

\end{thebibliography}
\end{document}